# Lightweight Convolutional Approaches to Reading Comprehension on SQuAD

**Tobin R. Bell**[*]
Stanford University
tbell@cs.stanford.edu

**Benjamin L. Penchas**[*]
Stanford University
bpenchas@cs.stanford.edu

## Abstract

Current state-of-the-art reading comprehension models rely heavily on recurrent neural networks. We explored an entirely different approach to question answering: a convolutional model. By their nature, these convolutional models are fast to train and capture local dependencies well, though they can struggle with longer-range dependencies and thus require augmentation to achieve comparable performance to RNN-based models. We conducted over two dozen controlled experiments with convolutional models and various kernel/attention/ regularization schemes to determine the precise performance gains of each strategy, while maintaining a focus on speed. We ultimately ensembled three models: *crossconv* (0.5398 dev F1), *attnconv* (0.5665), and *maybeconv* (0.5285). The ensembled model was able to achieve a 0.6238 F1 score using the official SQuAD evaluation script. Our individual convolutional model *crossconv* was able to exceed the performance of the RNN-plus-attention baseline by 25% while training 6 times faster.

## 1 Introduction

Stanford Question Answering Dataset (SQuAD) is a reading comprehension dataset introduced by Rajpurkar et al. (2016). It contains over 100,000 question-answer pairs on over 500 Wikipedia articles. Each question-answer pair contains a question of roughly 30 words with a correct answer that is a span of text from the corresponding reading passage. For example, take the following question-answer pairs constructed from Nikola Tesla's Wikipedia page:

> Nikola Tesla (Serbian Cyrillic: Никола Тесла; 10 July 1856–7 January 1943) was a Serbian American inventor, electrical engineer, mechanical engineer, physicist, and futurist who is best known for his contributions to the design of the modern alternating current (AC) electricity supply system.
>
> **In what year was Nikola Tesla born?**
> Ground truth: 1856 1856 1856
>
> **What was Nikola Tesla's ethnicity?**
> Ground truth: Serbian Serbian Serbian

Figure 1: Passage about Nikola Tesla, and questions with ground-truth answers.

Current state-of-the-art models, as judged by the SQuAD leaderboard, are predominantly RNNs with some unique attention mechanism. While each of these models has a distinctive approach to attention, at their core they are all RNNs with LSTM/GRU cells.

---

[*] Both authors contributed equally.

In contrast to these architectures, we wanted to explore convolutions applied to text. By their nature, convolutions are fast to train and infer since, unlike RNNs, they process tokens in parallel.

## 2  Approach and related work

The inspiration for a convolutional approach without RNNs came from Wei Yu et al. [2]. By using convolutions instead of RNNs, Wei Yu et al. were able to process tokens in parallel and train much faster. Their final model was able to achieve accuracy on par with recurrent methods while training 13 times faster. The convolutional method outlined in Wei Yu et al. [2] relies heavily on modular encoder blocks that employ positional encoding, convolutional layers, layer norm, and residual connections.

We quickly realized, however, that we would not be able to fully emulate Wei Yu et al. [2]. Even a drastically simplified version of their encoder block includes ~700,000 parameters, and their model uses 8 of these blocks. Combining that with the large memory demands of attention mechanisms, even the most capacious Azure GPUs we could use ran out of memory once we initialized a few of these encoder blocks. We thus focused instead on simply attempting to outperform the provided RNN baseline with a lightweight convolutional model.

At a high level, solving the reading comprehension task requires an understanding of the context, the question, and how they relate. Our approach was to use convolutions to capture local dependencies and attention to capture global dependencies as well as how the question relates to the context. To precisely evaluate what approaches to a convolutional model would produce the best results, we conducted successive controlled experiments to assess the value of any given change. By changing only one aspect of our model between most experiments, we gained a granular understanding of what worked well. Ultimately, this allowed us to selectively combine our most promising modifications to produce several successful lightweight convolutional models.

## 3  Experiments

In total, we trained 26 different models, though a few of them failed to provide us with useful conclusions due to technical shortcomings (mostly limited GPU memory) and scientific errors on our part (being unable to debug failed models due to changing too many variables at once). Performance measures of a selection of our most illuminating models are given below. EM and F1 scores were measured on the SQuAD dev set. Train time is the amount of time that passed during training before each model achieved its highest F1 score. For further details, summaries of all 19 experiments that yielded informative results are given in Table 3 at the end of this paper.

Table 1: Performance of selected models.

| Name | EM | F1 | Train Time |
|---|---|---|---|
| *baseline** | 0.2930 | 0.4007 | 6h 15m |
| *tpu100** | 0.2955 | 0.4061 | 6h 35m |
| *simpconv* | 0.1614 | 0.2333 | 33m |
| *triconv* | 0.1935 | 0.2740 | 1h 2m |
| *windowconv100* | 0.2075 | 0.2922 | 1h 18m |
| *narrowconv* | 0.2038 | 0.2822 | 1h 44m |
| *shareconv* | 0.2815 | 0.3922 | 1h 46m |
| *combconv100* | 0.3721 | 0.5114 | 1h 11m |
| *maybeconv* | 0.3912 | 0.5285 | 1h 58m |
| *crossconv* | 0.3990 | 0.5398 | 1h 34m |

* denotes an RNN-based model.

From these cursory performance metrics, it is clear that convolutional models train faster than RNN-based models. Even with significant augmentation, our convolutional models achieve their highest F1 score roughly 4 to 5 times faster than the baseline RNN model, which itself constitutes only a simple implementation of such a model (improved RNN models would take even longer to train). In the case of our later models, the achieved F1 score also exceeds the given RNN model's performance by roughly 25%.

It is also clear that convolutional models tend to perform worse than RNN-based models on the SQuAD task. For example, *simpconv*, our baseline-equivalent convolutional model (created by simply replacing the RNN encoder in *baseline* with two 4-layer CNNs), achieved a maximum F1 score only slightly greater than half of the baseline model's F1 score. Our experimentation thus focused on identifying specific enhancements that would allow convolutional models to compete with the given RNN-based approach. Our findings are described below.

### 3.1 Model encoding

Wei Yu et al. [2]'s model uses "input encoder" blocks to encode their question and context prior to applying attention between them. The given RNN *baseline* model and our simple convolutional *simpconv* both use a similar mechanism, in the form of a bidirectional LSTM and 4-layer CNN, respectively. However, Wei Yu et al. [2] also uses "model encoder" blocks after after context-to-question attention flow. In their model, these are fairly hefty: 3 applications of 7 blocks, each consisting of 2 convolutional layers and self-attention.

To keep our model lightweight, we used another instance of our generic encoder block (4-layer CNN) after applying context-to-question attention. This *triconv* model outperformed its predecessor *simpconv* model by 17.4%, achieving an F1 score of 0.2740 over *simpconv*'s score of 0.2333. At this point, it is worth noting that both models still lag behind the RNN-based *baseline*.

### 3.2 Wide kernel output layers

The *baseline* output layer works by simply applying a fully connected layer to the output of the context-to-question attention step. It then performs two projected softmax normalizations across the encoded context sequence to obtain two different distributions $p_{start}$ and $p_{end}$ (these are then argmax-ed for the prediction step). This offers no room for context-awareness when trying to label the bounds of a span, since all transformations happen at the level of individual words. We decided to replace this output layer with one that could account for information about neighboring words.

Our *windowconv100* model uses two convolutional layers to predict the start and end positions. These have a single filter (they must output only a single logit value) but their kernel width is 20, allowing them to look across a span of 20 words to evaluate how likely a given word is to start or end the true answer span. With such a wide kernel, in order to avoid overfitting and creating too many parameters, we precede this step with a fully connected layer (as does the baseline) that projects the hidden sequence representation down to 20 dimensions (from 200 in *windowconv100*, or 128 in our later models). Using this wide-windowed output layer achieved an F1 score of 0.2922, a ~6.6% improvement over the 0.2740 F1 of its parent model, *triconv*.

Another small augmentation we made to our output layer in our final models was to constrain the end position to come after the start position at prediction time. This replaced the naïve argmax method we had carried over from the baseline model into *windowconv100*, which predicted out-of-order span boundaries roughly 17% of the time.

### 3.3 Avoiding overfitting

Our convolutional models consistently achieved F1 scores that were roughly 0.2 higher on the train set than on the dev set. This consistent discrepancy indicated overfitting, and we attempted various methods of regularization to address it. Standard mechanisms like dropout and L2 regularization closed the gap between train and dev performance, but at large cost to trainability. Even after augmenting our training with layer normalization (in the style of Ba et al. [3]) and various optimizer tweaks, using these regularization strategies resulted in training times greater than the baseline RNN model, which did not align with our goals of finding a lightweight, fast-to-train model that performed well. However, two methods did work well, and those were reducing the width of the kernel in our convolutional layers, and reusing weights between the question and context input encoders. These improvements are described in more detail below.

### 3.3.1 Kernel narrowing

Though Wei Yu et al. [2] used 7-wide kernels in their ultra-deep CNN architecture, they also used various strategies to increase the diversity of their training data. Since we were not focused on augmenting our training data, we found that decreasing the kernel with to 5 and ultimately to 3 increased our dev/F1 and dev/EM scores by making it more difficult for the model to simply memorize N-grams from the training data. Our *narrowconv* model achieved an F1 of 0.2822 over its ancestor model, *triconv*, which achieved only 0.2740. This was also accompanied by a ~10% reduction in model size (a welcome bonus). We believe that smaller windows did not hurt the network's ability to process local dependencies because the use of multiple stacked convolutional layers still allowed word information to flow several steps to the left or right between layers.

### 3.3.2 Sharing weights

Initially, we trained our *triconv* model to use two distinct convolutional input encoders—one for the question and one for the context. Our thinking was that since questions and statements are syntactically and semantically in opposition to one another, it would be beneficial to allow the model to learn to process them differently. In practice, however, it simply facilitated overfitting on each one. By using a single convolutional encoder for both the question and the context in our *shareconv* model, we not only reduced the size of our model by ~25%, we also significantly increased its dev/F1 score to 0.3922, from 0.2740 in *triconv*. After the result of this experiment, it became clear that maximizing the network's ability to generalize was crucial. Using a single input encoder accomplishes that by forcing the model to learn a single language model for contexts and questions. Any structural differences between "statements" and "questions" would then be learned in the deeper layers of the network. This was the first convolutional model we tried that achieved performance on par with the RNN *baseline* approach.

## 3.4 Self-attention

Since each convolutional layer operates on only a small local neighborhood, one should expect a purely convolutional model to fare poorly at recognizing longer-range dependencies within the context and question. To overcome this weakness, we implemented multi-head self-attention in our encoder blocks, as outlined in Vaswani et al. [1], as a way to bridge gaps between spatially disparate but semantically related words within a single context or question. We followed their approach exactly (with one exception, described below in 3.4.1). We used only 4 heads, instead of their 8, and each head projected to a space of dimensionality 32 instead of 64. We did this to both decrease training time and memory requirements and to avoid overfitting (see above). Adding self-attention to our *maybeconv* model produced a performance gain of ~3.3%, increasing F1 to 0.5285 from the 0.5114 F1 achieved by its parent model, *combconv100*.

### 3.4.1 A note on bypass connections

The way that we integrated self-attention into a given model strongly affected the model's ability to train. In our first model to attempt self-attention, *triconv_attn*, we made the mistake of treating self-attention's output as a monolithic transformation, similar to a fully connected or convolutional layer. Given the output matrix $X$ of our 4 convolutional layers, we computed the encoder output matrix $Y$ using multi-head attention as follows:

$$Y = \text{SELFATTENTION}(X).$$

This works very poorly, due to the behavior of self-attention on sequences containing semantically distinctive words. Since distinctive words are unlikely to be considered "similar" to other words in the sentence (or even to themselves, owing to self-attention's use of different projection matrices for each head's key and query vectors), its attention output will become very small or zero. This proves too destructive to semantic information during forward propagation for the network to be able to perform well. It also greatly slowed down training progress, since the low similarity scores of distinctive words create very small gradients. Once we discovered this, we realized that the majority of models using self-attention (including Wei Yu et al. [2]) applied it through a residual connection:

$$Y = \text{SELFATTENTION}(X) + Y,$$

which allows a full identity of the original convolutional output to pass through. This solved our gradients problem, but still seemed too diluting to the output to allow later stages of the model to perform well. We then switched to dense connections, inspired by the work of Huang et al. [4] as an alternative to ResNets residual connections. These concatenate the attention output with the convolutional output:

$$Y = [\text{SELFATTENTION}(X), X].$$

This served to both maintain enough detail from the CNN output during forward propagation and allow gradients to flow around the self-attention layer without diminishing. This approach enabled the performance gains found in our *maybeconv* model, mentioned above.

### 3.5 Bidirectional attention

Wei Yu et al. [2] also introduced us to bidirectional attention—combining context-to-question attention and question-to-context attention using a trilinear similarity function. This is an approach originally described in Seo et al. [5] that not only uses the context to attend to relevant locations in the question, but vice-versa, allowing the question to attend to relevant portions of the context. This question-attended version of the context is then re-attended by the original context to produce a final representation.

While self-attention used projections and a simple dot-product similarity metric, Seo et al. [5] uses no projections and a single trainable "trilinear" similarity function:

$$\text{SIMILARITY}(q, c) = w_0 [q, c, q \odot c].$$

Here $w_0$ is a learned vector and $\odot$ represents element-wise multiplication. The above trilinear function is used to compute a similarity matrix $S \in \mathbf{R}^{n \times m}$ between each pair of context (length $n$) and question (length $m$) words. Using the **softmax** function, they then separately normalize along the rows and columns of $S$, yielding $\overline{S}$ and $\overline{\overline{S}}$ respectively. Lastly they compute the context-to-question attention as $A = \overline{S}Q^T$ and the question-to-context attention as $B = \overline{S}\,\overline{\overline{S}}^T C^T$, where $Q$ and $C$ are the encoded question and context.

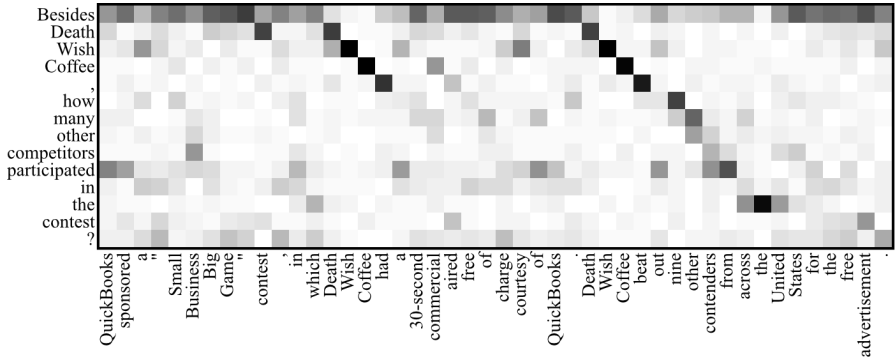

Figure 2: Visualized context-to-question attention weights.

Replacing the naive attention mechanism in the RNN baseline with this bidirectional mechanism achieved an F1 score of 0.5398 in *crossconv*, representing a ~5.6% improvement from the 0.5114 F1 achieved by its parent model, *combconv100*.

As seen in Figure 2, our final model, *attnconv*, which uses both self-attention and bidirectional cross-attention, learned to attend to relevant parts of the question when encoding the context. We see that "Death Wish Coffee" strongly attends to "Death Wish Coffee" which makes sense given the unique nature of that name. We see that "nine" in the context, which happens to be the correct answer span, attends to "how" and "many" which suggests that *attnconv* understands numbers to be good answers to questions like "how many." We also see "contenders" attends to "other," "competitors," and "participated" which suggests an understanding of the notion of competitors to Death Wish Coffee. Note that we see some blurring here (e.g. "contenders" attending equally to "competitors" and "participated") due to previously applied convolutional layers. A full structural diagram of *attnconv* is given in Figure 6.

## 4   Combining successful models

After our extensive experimentation, we combined features that we had found to work well. This allowed us to produce several successful models for our goals. For example, our *crossconv* model consists of features from *simpconv*, *triconv*, *narrowconv*, *windowconv*, *shareconv*, plus bidirectional attention (from Seo et al. [5]) and other extrapolative improvements. As can be seen in Figure 3, this combined *crossconv* model compounded the successes of its component features, allowing a purely convolutional approach to exceed the RNN baseline's performance while being smaller and faster to train and run.

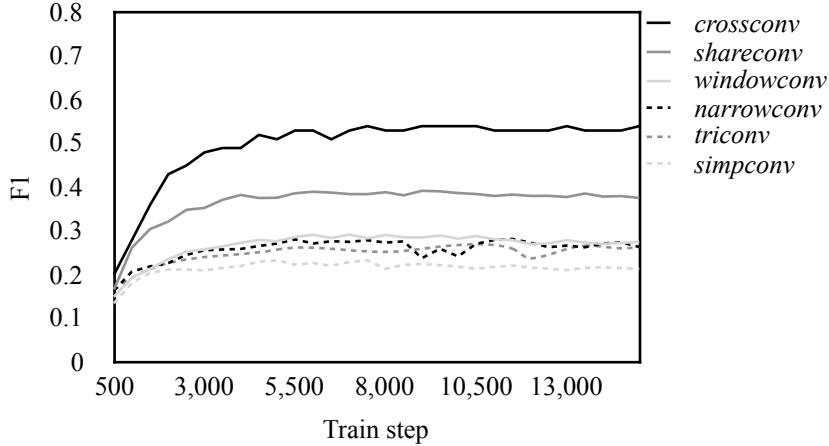

Figure 3: Effects of combining successful models into *crossconv*.

### 4.1   Ensemble methods

Table 2: Performance of individual and ensemble models.

| Name | EM | F1 |
|---|---|---|
| *attnconv* | 0.4494 | 0.5565 |
| *maybeconv* | 0.4628 | 0.5739 |
| *crossconv* | 0.4743 | 0.5835 |
| **ensemble** | **0.5195** | **0.6238** |

Ensemble learning improves prediction performance by combining several pre-trained models into one meta-model. To combine everything learned by the best models from Table 1, we chose to pursue ensemble methods that combine these models at prediction time. To do so, we created a confidence score that each model outputs with every prediction, defined as the product ($p_{start}$)($p_{end}$) for the chosen span. The ensemble model then chooses a final answer based on which model's prediction has the highest confidence, as shown in Figure 4. This strategy achieved our highest F1 score of 0.6238, outperforming all of its component models taken individually.

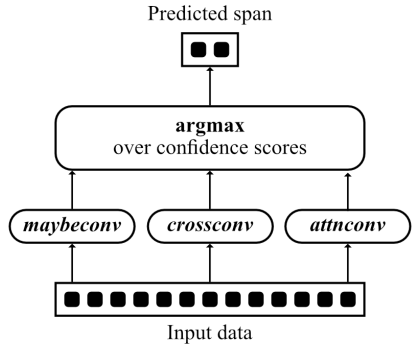

Figure 4: Ensemble combination of performant models.

## 4.2 Throughput

One focus of our research was building lightweight, fast models. The time complexity of prediction using our final *attnconv* model is $O((C + Q)H^2 + (C^2 + Q^2)H)$, where $C$ is the context length, $Q$ is the question length, and $H$ is the hidden size (note that the most significant slowdown here is the attention mechanism). As seen in Figure 5, our most performant models had few parameters and high throughput. We achieved our goal of producing lightweight, fast-to-train models that perform reasonably well.

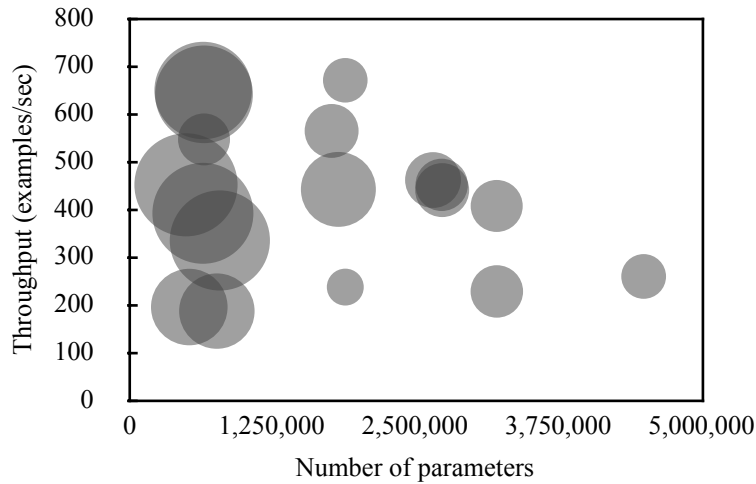

Figure 5: Time/space performance; diameter = F1 score.

## 5 Further work

To further improve the model, we believe incorporating character level embeddings would work well. Wei Yu et al. [2] successfully concatenated pre-trained word embeddings with the output of convolving over trainable character embeddings for the word's characters. We believe these trainable character embeddings would increase performance of the model without greatly slowing down training.

Wei Yu et al. [2] also successfully incorporated back-translation as a data augmentation strategy, and we believe this strategy would work well for our model; since our model converges very quickly, it could easily be trained on a much larger dataset. To augment the dataset, we would employ a translation model from English to some other language and back again, allowing us to get paraphrases of the question-answer pairs in the original dataset.

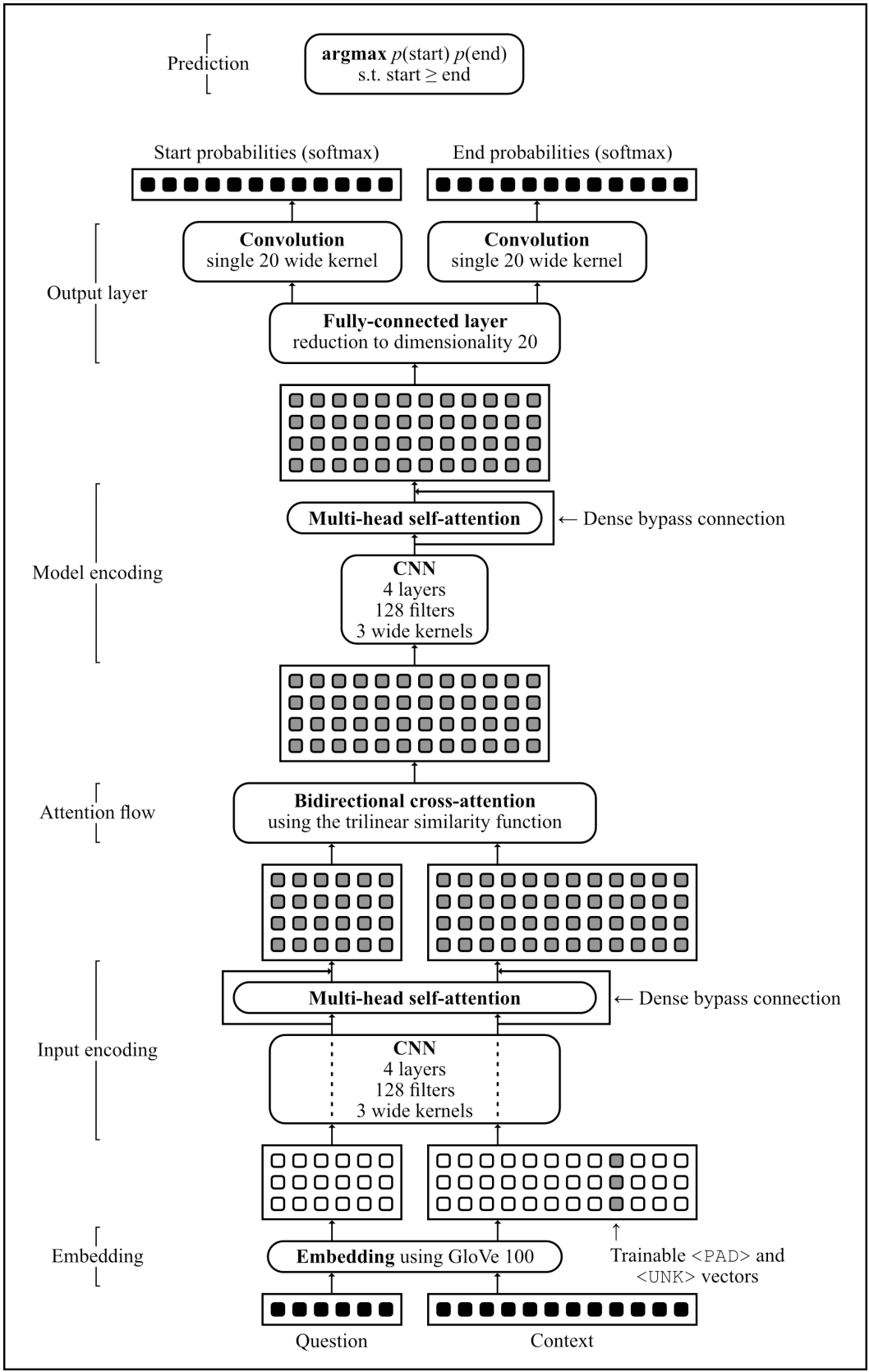

Figure 6: Detailed model diagram of *attnconv*.

# Supplementary material

Table 3: Summary of all informative experiments (*eps*: examples per second).

| Name / Description | Motivation | Results |
| --- | --- | --- |
| *baseline* \| Single bidirectional GRU input encoder, context-to-question attention, one F.C. model encoder layer. Constant GloVe 100. | Given model. | 0.4007 F1<br>521,802 params<br>195.7 eps |
| *tpu100* \| Derived from (D.F.) *baseline*. Constant GloVe 100 embeddings, with trainable `<PAD>` and `<UNK>` vectors (T.P.U.). | Remove noise introduced by constant random `<UNK>` and `<PAD>` vectors. | 0.4061 F1<br>522,002 params<br>195.1 eps |
| *tpu300* \| D.F. *tpu100*. Constant GloVe 300 embeddings instead of 100, with T.P.U. | Compare effects of large versus small embeddings on overfitting. | 0.3947 F1<br>762,402 params<br>187.2 eps |
| *simpconv* \| D.F. *baseline*. 2 CNNs (4 layers each, kernel width 5, ReLU), one each for the context and the question, instead of GRU. GloVe 300. | Establish a baseline for CNN-based performance. | 0.2333 F1<br>1,882,602 params<br>670.8 eps |
| *triconv* \| D.F. *simpconv*. Adds another CNN (4 layers, kernel width 5, ReLU) as a model encoder after context-to-question attention. GloVe 300. | Process context-question relationship after applying attention between them. | 0.2740 F1<br>2,723,402 params<br>451.7 eps |
| *triconv_attn* \| D.F. *triconv*. Multi-head self-attention (4 heads, head size 50, residual connections) after 2 conv layers in each encoder. | Allow encoders to resolve long-range dependencies using attention. | 0.1932 F1<br>1,882,602 params<br>237.5 eps |
| *triconv_reg* \| D.F. *triconv*. L2 loss on all trainable variables, and dropout after the final feed-forward layer of an encoder block. GloVe 300. | Reduce overfitting by regularizing conv kernels and dropping out. | 0.2723 F1<br>3,203,402 params<br>407.5 eps |
| *windowconv100* \| D.F. *triconv*. Replaces baseline output layer with two wide convolutional layers (kernel width 20) for start and end. GloVe 100. | Widen context-awareness when predicting start and end positions. | 0.2922 F1<br>2,647,822 params<br>461.8 eps |
| *attn2* \| D.F. *triconv_attn*. Self-attention after all 4 convolutional layers instead of between them. Uses layer norm before self-attention. GloVe 300. | Resolve backprop challenges from *triconv_attn*. | 0.2747 F1<br>3,204,602 params<br>228.4 eps |
| *shareconv* \| D.F. *triconv*. Shares parameters between the two input encoder blocks. | Reduce model size and overfitting by learning only one language model. | 0.3922 F1<br>1,822,402 params<br>442.2 eps |
| *windowconv300* \| D.F. *windowconv100*. GloVe 300 instead of 100. | Compare effects of large versus small embeddings on overfitting. | 0.2824 F1<br>2,727,822 params<br>440.7 eps |
| *narrowconv* \| D.F. *triconv*. Reduces kernel width from 5 to 3 in all convolutional layers. GloVe 300. | Discourage memorizing n-grams from the training set to reduce overfitting. | 0.2822 F1<br>1,763,402 params<br>564.7 eps |
| *combconv100* \| Merges successful models (*shareconv*, *windowconv100*, *narrowconv tpu100*). Hidden size 150. Constraint: *start* ≤ *end*. | Combine successes of previous convolutional models. | 0.5114 F1<br>650,322 params<br>641.4 eps |
| *combconv50* \| D.F. *combconv100*. GloVe 50 with T.P.U. | Reduce overfitting with smaller word vectors. | 0.5101 F1<br>642,722 params<br>649.4 eps |
| *dropoutconv* \| D.F. *combconv100*. GloVe 100 with T.P.U. Dropout (0.5) applied before every convolutional layer. | Reduce overfitting by regularizing the network with dropout. | 0.2721 F1<br>650,322 params<br>546.9 eps |

| Name / Description | Motivation | Results |
| --- | --- | --- |
| ***maybeconv*** \| D.F. *combconv100*. Hidden size 128. Multihead self-attention (4 heads, 32 channels each) with dense bypass connections. | Better resolve long-range dependencies using self-attention. | 0.5285 F1<br>640,566 params<br>392.1 eps |
| ***deepconv*** \| D.F. *maybeconv*. Adds two more encoder blocks that process blended representations produced by basic attention. | Increase the power of the model by going deeper, to better fit the train set. | 0.2342 F1<br>4,485,402 params<br>259.8 eps |
| ***crossconv*** \| D.F. *combconv100*. Reduces hidden size to 128. Uses bidirectional attention with trilinear similarity. GloVe 100 with T.P.U. | Increase capability of context-question attention. | 0.5398 F1<br>492,982 params<br>451.8 eps |
| ***attnconv*** \| D.F. *crossconv* and *maybeconv*. Both self-attention and bidirectional attention. 8 self-attention heads. GloVe 100 with T.P.U. | Combine both of our successful attention mechanisms. | 0.5242 F1<br>788,406 params<br>335.2 eps |